\documentclass[10pt,twocolumn,letterpaper]{article}

 \usepackage{cvpr}
\usepackage{times}
\usepackage{epsfig}
\usepackage{graphicx}
\usepackage{amsmath}
\usepackage{amssymb}
\usepackage{dsfont}
\usepackage{amssymb}
\usepackage{tipa}
\usepackage{mathrsfs}
\usepackage{textcomp}
\usepackage{tabularx}
\usepackage{subcaption}
\usepackage{adjustbox}
\usepackage{array,multirow,graphicx}
\usepackage{makecell}
\usepackage{boldline}
\usepackage{multicol}

\cvprfinalcopy
\newcommand*{\affaddr}[1]{#1} 
\newcommand*{\affmark}[1][*]{\textsuperscript{#1}}
\newcommand*{\email}[1]{\texttt{#1}}

\graphicspath{{./images/}}

\usepackage[pagebackref=true,breaklinks=true,letterpaper=true,colorlinks,bookmarks=false]{hyperref}

\begin{document}


\title{VisDA: The Visual Domain Adaptation Challenge}

\author{
 Xingchao Peng\affmark[1], Ben Usman\affmark[1], Neela Kaushik\affmark[1], Judy Hoffman\affmark[2], Dequan Wang\affmark[2] and Kate Saenko\affmark[1]\\
 {\small \email{xpeng,usmn,nkaushik,saenko@bu.edu}}, {\small \email{jhoffman,dqwang@eecs.berkeley.edu}}\\
 \affaddr{\affmark[1]Department of Computer Science, Boston University}\\
 \affaddr{\affmark[2]EECS, University of California Berkeley}
 }

\maketitle

\begin{abstract}
    We present the 2017 Visual Domain Adaptation (VisDA) dataset and challenge, a large-scale testbed for unsupervised domain adaptation across visual domains.  
    Unsupervised domain adaptation aims to solve the real-world problem of domain shift, where machine learning models trained on one domain must be transferred and adapted to a novel visual domain without additional supervision. The VisDA2017 challenge is focused on the simulation-to-reality shift and has two associated tasks: image classification and image segmentation. The goal in both tracks is to first train a model on simulated, synthetic data in the source domain and then adapt it to perform well on real image data in the unlabeled test domain. Our dataset is the largest one to date for cross-domain object classification, with over 280K images across 12 categories in the combined training, validation and testing domains. The image segmentation dataset is also large-scale with over 30K images across 18 categories in the three domains. We compare VisDA to existing cross-domain adaptation datasets and provide a baseline performance analysis, as well as results of the challenge. 
\end{abstract}

\section{Introduction}

\begin{figure}[t]
    \centering
    \includegraphics[width=\linewidth]{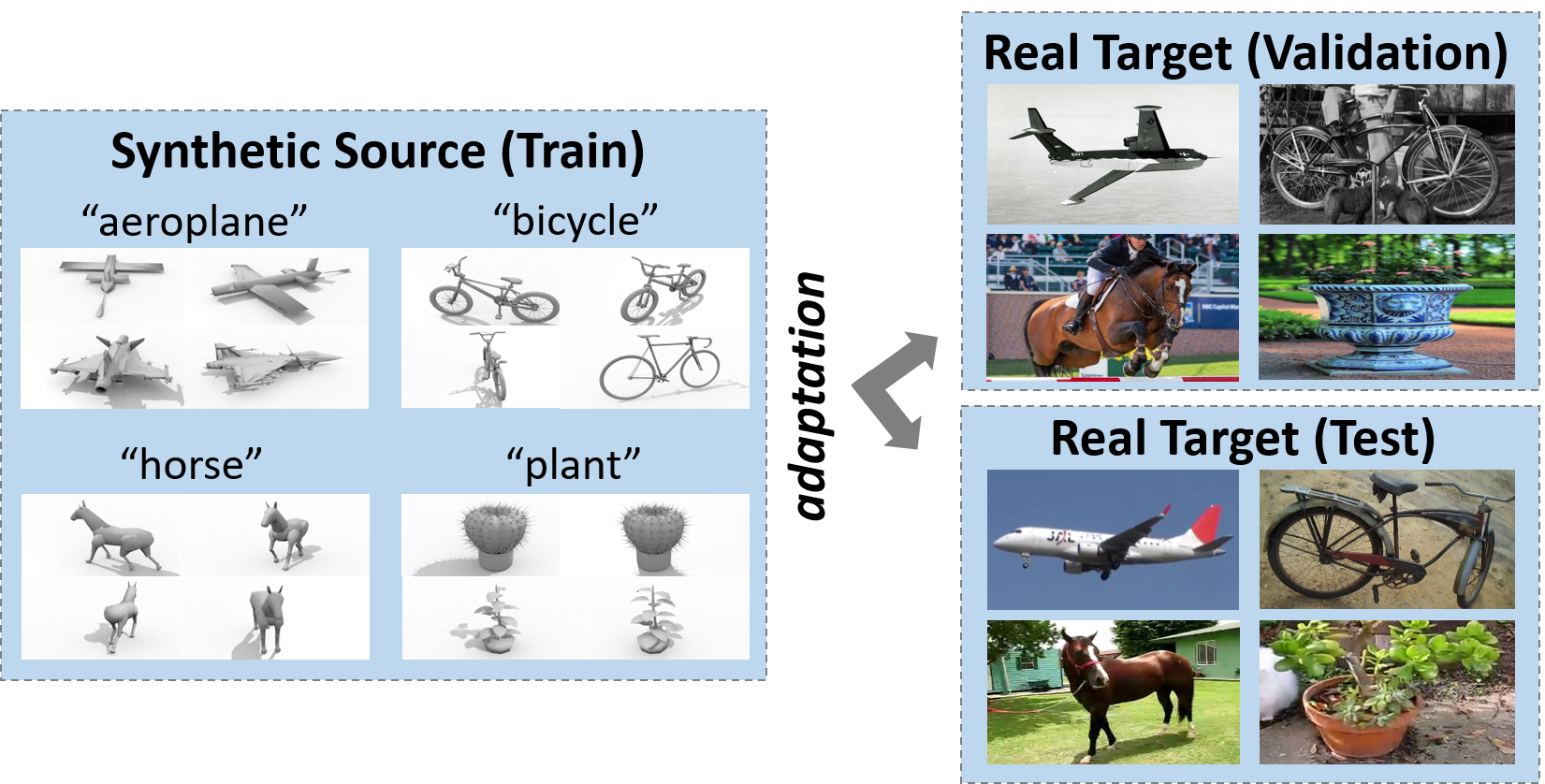}
        {\small (a) Image Classification Task}\vspace{2mm}  
   
    \includegraphics[width=\linewidth]{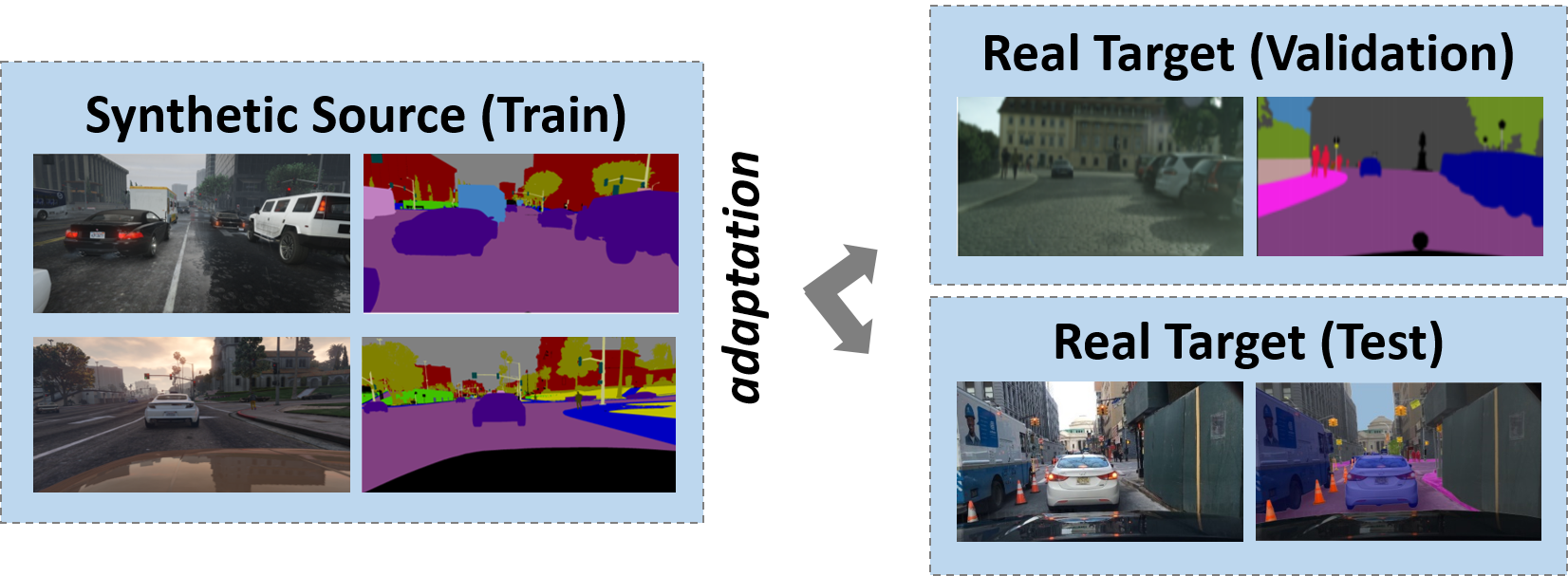}
        {\small (b) Semantic Image Segmentation Task}
    \caption{\small  (Best viewed in color) The VisDA2017 challenge aims to test models ability to perform \textit{unsupervised domain adaptation}, i.e. to transfer  knowledge from a large labeled source domain to an unlabeled target domain. It contains a challenging simulation-to-real domain shift and consists of two tasks: (a) classification and (b) semantic segmentation. For each task we provide data from \textit{three~distinct} domains: train (source), validation (target) and test (target), therefore challenging domain adaptation methods' ability to perform well out-of-the-box on unseen domains without manual hyper-parameters tuning. 
    }
    \label{fig:overview}
    \vspace{-0.6cm}
    \centering
    
\end{figure}

It is well known that the success of machine learning methods on visual recognition tasks is highly dependent on access to large labeled datasets. Unfortunately, model performance often drops significantly on data from a new deployment domain, a problem known as \textit{dataset shift}, or \textit{dataset bias}~\cite{datashift_book2009}. Changes in the visual domain can include lighting, camera pose and background variation, as well as general changes in how the image data is collected. While this problem has been studied extensively in the \textit{domain adaptation} literature~\cite{Csurka17}, progress has been  limited by the lack of large-scale challenge benchmarks. 

Challenge competitions drive innovation in computer vision, however, most challenges build their train and test splits by randomly sampling data from the same domain. An assumption of test data having same distribution as train data rarely holds in real-life deployment scenarios. Cross-domain benchmarks~\cite{nene1996columbia, office, carlucci2017autodial, Tommasi14} have been created for evaluating domain adaptation algorithms, but they often have low task diversity, small domain shifts and small dataset size.

This paper introduces the \textit{Visual Domain Adaptation (VisDA)} dataset and a challenge of the same name to encourage further progress in developing more robust domain transfer methods. The challenge task is to train a model on a ``source" domain and then update it so that its performance improves on a new ``target" domain, without using any target annotations. Our challenge includes two tracks, illustrated in Figure~\ref{fig:overview}. The first track introduces the more traditional adaptation for \textit{object classification} task. The second one suggests to tackle the relatively less-studied \textit{semantic segmentation} task, which involves labeling each pixel of an input image with a semantic label. 

The VisDA2017 challenge focuses on the domain shift from simulated to real imagery--a challenging shift that has many practical applications in computer vision. This type of ``synthetic-to-real" domain shift is important in many real-world situations when labeled imagery is difficult or expensive to collect (autonomous decision making in robotics, medical imaging, etc.), whereas synthetic rendering pipeline can produce virtually infinite amounts of labeled data ones set up. For this reason we generated the largest cross-domain synthetic-to-real object classification dataset to date with over 280K images in the combined training, validation and testing sets. For the semantic segmentation track we augmented existing datasets for a total of approximately 30k images across three domains.

The VisDA2017 challenge focuses on \textit{unsupervised} domain adaptation (UDA) for which the deployment domain images are not labeled. 
For each task, we provide labeled samples from training domain (source) and unlabeled samples from validation (target) and a different test (target) domains. While there may be scenarios where target labels are available (so called ``supervised domain adaptation"), we focus on purely unsupervised adaptation case as it is more challenging and often more realistic.
Many unsupervised domain adaptation evaluation protocols use some labeled target domain data to select hyperparameters. However, assuming labeled target data goes against the UDA problem statement. For this reason, we collect two \textit{different} target domains, one for  validation of hyperparameters and one for testing the models.

In this paper, we present both our baseline results and most interesting results obtained by participants during the actual challenge. 
In addition to our multi-domain datasets released during the main timeline of the challenge, we plan to fully opensource and support our model collection along with scripts and metadata required for proper rendering, and propose a more challenging version of the classification dataset as well as present multiple more challenging experiment setups - all in order to lay the foundation for future research on this topic. 
\section{Related Work}

There has been a lot of prior work on visual domain adaptation, ranging from the earlier shallow feature methods~\cite{office,BergamoTorresani10,duan10} to the more recent deep adaptation approaches~\cite{ganin2014unsupervised,Tzeng_2015_ICCV}.
A review of existing work in this area is beyond the scope of this paper; we refer the reader to a recent comprehensive survey~\cite{Csurka17}. 
Several benchmark datasets have been collected and used to evaluate visual domain adaptation, most notable are summarized in Table~\ref{tab:datasets}. Majority of popular benchmarks lack  task diversity: the most common cross-domain datasets focus on the image classification task, i.e. digits of different styles, objects~\cite{office} or faces~\cite{sim2002cmu} under varying conditions. 
Other tasks such as detection~\cite{peng2015learning}, structure prediction \cite{RosCVPR16, Cordts2016Cityscapes, richter2016playing} and sequence labeling \cite{graves2012supervised} have been relatively overlooked.

\noindent\textbf{Classification Datasets.} Adaptation of image classification methods has been among the most extensively studied problems of visual domain adaptation. One of the difficulties in re-using existing datasets to create multi-domain benchmarks is that the same set of categories must be shared among all domains. For digits (ten categories, 0-9), the most popular benchmark setup consists of three domains: MNIST (handwritten digits)~\cite{lecun1998gradient}, USPS (handwritten digits)~\cite{hull1994database} and SVHN (street view house numbers)~\cite{netzer2011reading}. Digit images are sometimes synthetically augmented to create additional domains, for example by inverting colors or using randomly chosen backgrounds \cite{ganin2014unsupervised}.

The Office dataset~\cite{office} is a popular benchmark for real-world objects. It contains 31 object categories captured in three domains: office environment images taken with a high quality camera (DSLR), the same environment captured with a low quality webcam (WEBCAM), and images downloaded from the amazon.com website (AMAZON). 

Besides the small size of these benchmarks, another problem is the relatively small domain shifts, such as the shift between two different sensors (DSLR vs Webcam in the Office dataset~\cite{office}), or between two very similar handwritten digit datasets (MNIST vs USPS). Over time, improvements in underlying image representations and adaptation techniques have closed the domain gap on these benchmarks, and more challenging domain shifts are now needed to drive further progress.
Another issue is the small scale. Modern computer vision methods require a lot of training data, while cross-domain datasets such as Office Dataset~\cite{office} only contain several hundred of images. The Cross-Dataset Testbed~\cite{Tommasi14} is a more recent classification benchmark. its ``dense" version contains 40 classes extracted from Caltech256, Bing, SUN, and Imagenet with a minimum of 20 images per class in each dataset. It is significantly larger than Office, however, some domains are fairly close as they were collected in a similar way from web search engines. On the Caltech-Imagenet shift, adaptation performance has reached close to 90\% accuracy~\cite{carlucci2017autodial}.

The popularity of the image classification task as a testbed may be due to the relative simplicity of the task and the lower effort required to engineer a good baseline model. Compared with other vision problems such as object detection, activity detection in video, or structured prediction, image classification is simpler and less computationally expensive to explore. Moreover, many state-of-the-art classification models are readily available for use as a baseline upon which adaptation can be applied. At the same time, other tasks have characteristics that could present unique challenges for domain adaptation. In this work, we propose experimental setups for both the more common classification task, and the less studied semantic segmentation task.

\begin{table}[t]
    \centering 
    OBJECT CLASSIFICATION \\
    \begin{tabularx}{\linewidth}{|c|ccc|} \hline
    Dataset & Examples & Classes & Domains \\ \hline
    COIL20 \cite{nene1996columbia} & 1,440 & 20 & 1 (tools) \\
    Office \cite{office} & 1,410  & 31 & 3 (office) \\
    Caltech \cite{carlucci2017autodial}& 1,123 & 10 & 1 (office) \\
    CAD-office \cite{peng2015learning} & 775 & 20 & 1 (office) \\
    Cross-Dataset \cite{Tommasi14} & 70,000+ & 40 & 12 (mixed) \\ \hline
    \textbf{VisDA-C}   & 280,157 & 12 & 3 (mixed) \\ \hline
    \end{tabularx} 
~~~~~\\    
    SEMANTIC SEGMENTATION \\
    \begin{tabularx}{\linewidth}{|c|ccc|} \hline
    Dataset & Examples & Classes & \ Domains \ \\ \hline
    {\footnotesize SYNTHIA-subset} \cite{RosCVPR16} &  9,400 & 12 & 1  (city)\\ 
    CityScapes \cite{Cordts2016Cityscapes} & 5,000 & 34 & 1  (city)\\
    GTA5 \cite{richter2016playing} & 24,966 & 18 & 1  (city)\\ \hline
    \textbf{VisDA-S}   & 31,466 & 18 & 3 (city) \\    \hline
    \end{tabularx}
    \caption{\small Comparison of VisDA to existing cross-domain datasets used for domain adaptation experiments, with corresponding numbers of classes and domains. Datasets that share object categories can be combined to form cross-domain benchmarks. Table \ref{fig:datasets_extra} (supplementary) lists sizes of other cross-domain datasets used for different tasks: face recognition, object detection, digit classification. }

    \label{tab:datasets}
    \vspace{-0.5cm}
\end{table}

\noindent\textbf{Semantic Segmentation Datasets.} 
Semantic segmentation methods assign object labels to each pixel of an input image. Dataset annotation is a highly labor-intensive process, this is why there are only very few semantic segmentation datasets designed specifically for domain adaptation. Two datasets that are frequently paired together for visual segmentation tasks are SYNTHIA \cite{RosCVPR16} and CityScapes \cite{Cordts2016Cityscapes}. SYNTHIA provides a collection of synthetically generated urban images that mimic car dash-cam footage, while CityScapes provides images of real urban street scenes.  Other synthetic street-view datasets include GTA5 \cite{richter2016playing} and Virtual KITTI \cite{gaidon2016virtual}. Our VisDA benchmark combines GTA5, CityScapes, and data from the recently released Berkeley Deep Drive/Nexar dataset~\cite{nexet}.

\noindent\textbf{Synthetic Datasets.} Synthetic data augmentation has been extensively employed in computer vision research. More specifically, 3D models have been utilized to generate synthetic images with variable object poses, textures, and backgrounds~\cite{peng2015learning}. Recent usage of 3D simulation has been extended to multiple vision tasks such as object detection~\cite{peng2015learning, sun2014virtual}, pose estimation~\cite{su2015render}, robotic simulation~\cite{tzeng2015towards}, semantic segmentation~\cite{richter2016playing}. Popular 3D model databases of common objects that may be used in visual domain adaptation tasks include  ObjectNet3D \cite{xiang2016objectnet3d}, ShapeNet and the related ModelNet \cite{shapenet2015}. Table \ref{tab:synth_datasets} compares the VisDA classification dataset to existing synthetic object datasets. 

\begin{table}[t]
    \centering
    SYNTHETIC OBJECTS \\
    \begin{tabular}{|c|c c c|} \hline
    Dataset & Models & Images & Classes \\ \hline
    ModelNet \cite{wu20153d} & 127,915 & - & 662 \\
    PASCAL3D+ \cite{xiang2014beyond} & 77 & 30,899 & 12 \\
    ObjectNet3D \cite{xiang2016objectnet3d}& 44,147 & - & 100 \\
    ShapeNet-Core\cite{shapenet2015}& 51,300 & - & 55 \\
    ShapeNet-Sem\cite{shapenet2015}& 12,000 & - & 270 \\
    Redwood\cite{Choi2016}& 10,000 & - & 44 \\
    IKEA \cite{lpt2013ikea} & 219 & - & 11  \\ \hline
    \textbf{VisDA-C (source)}   & 1,907 &  152,397 & 12 \\ \hline
    \end{tabular}
\caption{\small Comparison between VisDA-C and existing synthetic object datasets. Majority of 3D model databases have thousands of rare classes with significantly unequal number of samples that are not present in the majority of other datasets that makes it difficult to use them for cross-domain adaptation. At the same time, VisDA-C dataset delivers a substantial balanced collection of models with all metadata (orientation, scaling, etc.) required for proper rendering and a deliberately limited number of classes mostly overlapping with ``standard" VOC PASCAL classes. }
\label{tab:synth_datasets}
\vspace{-0.5cm}
\end{table} 
\section{VisDA-C: Classification Dataset}

The \textit{VisDA Classification (VisDA-C)} dataset provides a large-scale testbed for studying unsupervised domain adaptation in image classification. The dataset contains three splits (domains), each with the same 12 object categories:
\vspace{-2mm}
\begin{itemize}
\item \textbf{training domain (source):} synthetic renderings of 3D models from different angles and with different lighting conditions,
\vspace{-2mm}
\item \textbf{validation domain (target):} a real-image domain consisting of images cropped from the Microsoft COCO dataset~\cite{mscoco14},
\vspace{-2mm}
\item \textbf{testing domain (target):} a real-image domain consisting of images cropped from the Youtube Bounding Box dataset~\cite{real17youtube} 
\vspace{-1mm}

\end{itemize}

We use different target domains for the validation and test splits to prevent hyper-parameter tuning on the test data. Unsupervised domain adaptation is usually done in a \textit{transductive} manner, meaning that unlabeled test data is actively used to train the model. However, it is not possible to tune hyper-parameter on the test data, since it has no labels. Despite this fact, the lack of established validation sets often  leads to poor experimental protocols where the labeled test set is used for this purpose. In our benchmark, we provide a validation set to mimic the more realistic deployment scenario where the target domain is unknown at training time and test labels are not available for hyper-parameter tuning. This setup also discourages algorithms that are designed to handle a specific target domain. It is important to mention that the validation and test sets are \textit{different} domains, so over-tuning to one can potentially degrade performance on another.

\begin{table}[t]
\begin{center}

\includegraphics[width=0.8\linewidth]{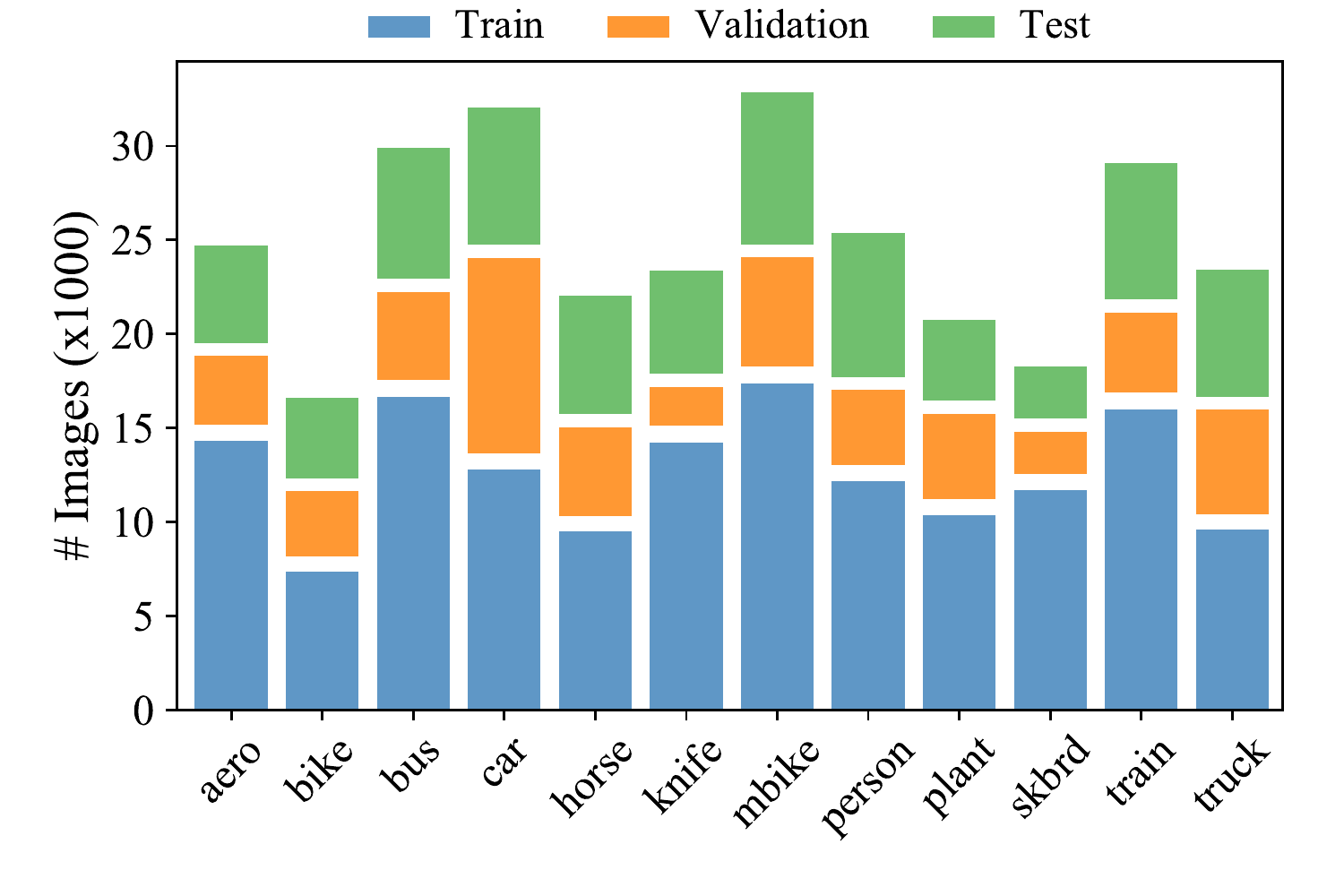}
\end{center}
\vspace{-0.9cm}
\caption{ \small Number of images per category in VisDA Classification training, validation and testing domains. Please refer to Table \ref{tab_num_of_models_and_images} (supplementary) for more details. }
\vspace{-0.6cm}
\label{pie_num_of_models_and_images}
\end{table}

\subsection{Dataset Acquisition}

\noindent \textbf{Training Domain: CAD-Synthetic Images}. The synthetic dataset was generated by rendering 3D models of the same object categories as in the real data from different angles and under different lighting conditions. As shown in Table~\ref{tab_num_of_models_and_images}, we obtained 1,907 models in total and generated 152,397 synthetic images. We used four main sources of models that are indicated with a \textit{sec} prefix of the corresponding image filename. These four sources include manually chosen subsets of ShapenetCore \cite{shapenet2015}, NTU 3D \cite{ntu3d}, SHREC 2010 \cite{shrec10} with some labels retrieved from TSB \cite{tsb} and our own collection of 3D CAD models from 3D Warehouse SketchUp. 

\begin{figure}[t]
\vspace{-0.5cm}
    \centering
    \includegraphics[width=\linewidth]{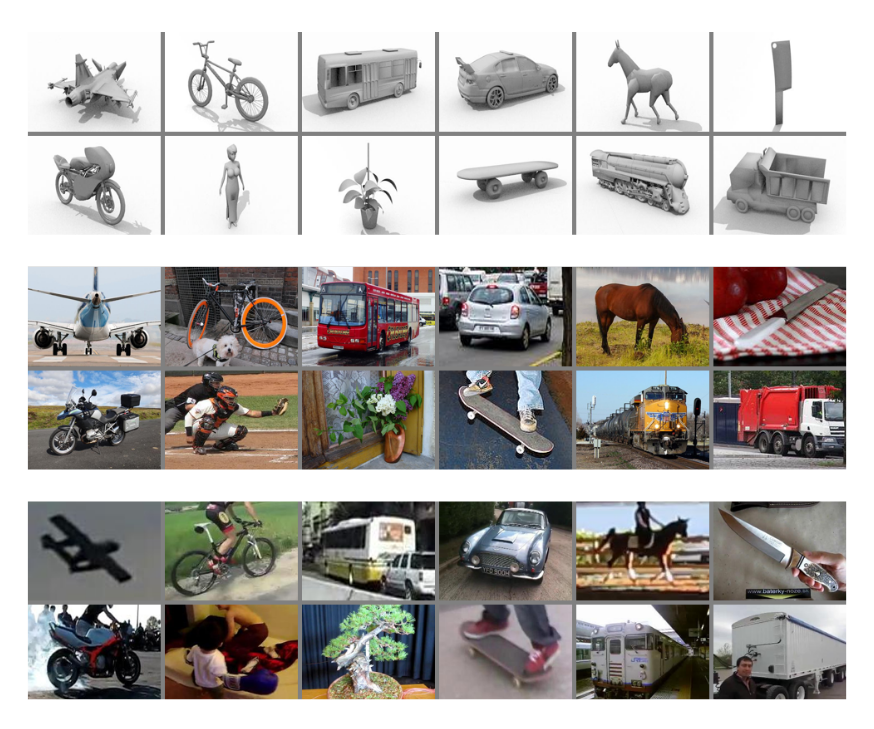}
    \vspace{-0.5cm}
    \caption{\small Sample images from the VISDA-C dataset. The top group shows synthetically rendered images (source domain), the middle group shows objects cropped from COCO dataset~\cite{mscoco14} using their bounding boxes (validation target domain), and the bottom group shows similarly cropped images from YouTube-BB dataset~\cite{real17youtube} (test target domain).}
    \label{fig_cls_sample}
    \centering
    \vspace{-0.5cm}
\end{figure}

We used twenty different camera yaw and pitch combinations with four different light directions per model. The lighting setup consists of ambient and sun light sources in 1:3 proportion. Objects were rotated, scaled and translated to match the floor plane, duplicate faces and vertices were removed, and the camera was automatically positioned to capture the entire object with a margin around it. For textured models, we also rendered their un-textured versions with a plain grey albedo. In total, we generated 152,397 synthetic images to form the synthetic source domain. Per-category image numbers are shown in Table~\ref{tab_num_of_models_and_images} (supplementary), and Figure \ref{fig_cls_sample} shows samples of training domain data. 

\begin{table*}[ht]
\centering
\vspace{-0.3cm}
\footnotesize{
\textit{\textbf{Training Domain (CAD-synthetic) $\rightarrow$ Validation Domain (MS COCO)}}
\noindent
\begin{tabular}{p{2.2cm}|p{0.4cm}p{0.37cm}|p{0.37cm} p{0.37cm} p{0.37cm} p{0.37cm} p{0.37cm} p{0.37cm} p{0.37cm} p{0.37cm} p{0.37cm} p{0.37cm} p{0.37cm} p{0.45cm}  !{\vrule width0.8pt} p{0.6cm} |p{0.8cm} | p{0.9cm}   }
\Xhline{0.8pt}
 Method   &Train &Test & aero  & bike & bus & car & horse & knife & mbike & person & plant & skbrd & train & truck  & Mean & Source & Gain\\
 \Xhline{0.8pt} 

DAN~\cite{long2015learning}&syn&real& 71.0 &	47.4 & 67.3 & 31.9 & 61.4 &	49.9 & 72.1 & 36.1 & 64.7 &	28 & 70.6 & 19 & 51.62 & 28.12 &  83.6\%\\
\footnotesize{{D-CORAL}}~\cite{SunS16a}&syn&real&76.5 & 31.8 & 60.2 & 35.3 &	45.7 & 48.4 & 55 & 28.9 & 56.4 & 28.2 & 60.9 & 19.1 & 45.53 & 28.12& 61.91\% \\
\hline
Source (AlexNet) & syn & real & 53.5 & 3.7 & 50.1 & 52.2 & 27.9 & 14.9 & 27.6 & 2.9 & 25.8 & 10.5 & 64.4 & 3.9 & 28.12 & - & - \\
Oracle (AlexNet) & syn & syn & 100 & 100 & 99.8 & 99.9 & 100 & 99.9 & 99.8 & 100 & 100 & 100	& 99.9 & 99.7 & 99.92 &-&- \\
Oracle (AlexNet) &real& real& 94.9& 83.2 & 83.1& 86.5& 93.9 & 91.8 &	90.9 & 86.6 & 94.9 &	88.9 & 87 &	65.4 & 87.26&-&- \\
\Xhline{0.8pt}
\end{tabular}

\vspace{0.1cm}
\textit{\textbf{Training Domain (CAD-synthetic) $\rightarrow$ Testing Domain (YT-BB)}}
\noindent\begin{tabular}{p{2.2cm}|p{0.4cm} p{0.37cm}|p{0.37cm} p{0.37cm} p{0.37cm} p{0.37cm} p{0.37cm} p{0.37cm} p{0.37cm} p{0.37cm} p{0.37cm} p{0.37cm} p{0.37cm} p{0.45cm}  !{\vrule width0.8pt} p{0.6cm} |p{0.8cm} | p{0.9cm}   }
\Xhline{0.8pt}
 Method   & Train & Test & aero  & bike & bus & car & horse & knife & mbike & person & plant & skbrd & train & truck  & Mean & Source & Gain \\
 
\Xhline{0.8pt}
 
DAN&syn&real &55.4 & 18.4 & 59.9 & 68.6 &55.2 & 41.4 & 63.4 & 30.3 & 78.8 & 23.0 & 62.8 & 40.1 & 49.78 & 30.81 & 61.57\%\\
D-CORAL& syn&real& 62.5 & 21.7 & 66.4 & 64.7 & 31.1 & 36.6 & 54.3 & 24.9 &	73.8 & 30.0 & 43.4 & 34.1 & 45.29 & 30.81 & 47.0\%\\
\hline
\noindent GF\_ColourLab\_UEA & syn & real &96.9	&92.4&	92.0&	97.2&	95.2&	98.8&	86.3&	75.3&	97.7&	93.3&	94.5&	93.3&	\textbf{92.8} & 45.3 & \textbf{104.8\% }\\
NLE\_DA  &syn & real &	94.3	&86.5&	86.9&	95.1&	91.1&	90.0&	82.1&	77.9&	96.4&	77.2&	86.6&	88.0&	87.7 & 64.3 & 36.4\%\\
BUPT\_OVERFIT &syn & real &	95.7&	67.0&	93.4&	97.2&	90.6&	86.9&	92.0&	74.2&	96.3&	66.9&	95.2&	69.2&	85.4 & 63.2 & 35.12\% \\
\hline
Source (AlexNet) &syn& real& 46.5 & 0.8 & 59.2 & 82.7 & 21.0 & 14.4 & 23.2 & 1.0 & 46.1 & 17.2 & 47.8 & 9.8 & 30.81 & - & - \\
Oracle (AlexNet) & real & real & 94.5 & 84.4 & 90.1 & 95.5 & 93.2 & 95.1 & 90.4 & 90.1 & 95.7 & 89.5 & 94.6 & 91.8 &  92.08 & - & - \\
Oracle (ResNext) & real & real & 96.2 & 89.3 & 92.8 & 98.3 & 94.8 & 95.7 & 90.7 & 92.0 & 95.9 & 86.0 & 94.9 & 93.5 & 93.40 & - & - \\

\Xhline{0.8pt}

\end{tabular}
\vspace{-0.05cm}
}
\caption{\small \textbf{Baseline and challenge results for the classification track}.
We show per-category accuracy for models trained using various adaptation methods. The top table reports performance of models adapted to \textit{validation} domain of VisDA-C. The bottom one reports adaptation performance for to the \textit{test} domain. First column indicates either the method used for adaptation, or a special setup (a source only case with no adaptation; a classifier trained using oracle label values). 
The tables show domain adaptation algorithms (DAN~\cite{long2015learning} and D-CORAL~\cite{SunS16a}) can improve the results by roughly 20 percent. Gain column indicates relative improvement over source model.}

\label{tab_cls}
\vspace{-0.5cm}
\end{table*}

\noindent \textbf{Validation Domain: MS COCO.} The validation dataset for the classification track is built from the Microsoft COCO \cite{mscoco14} \textit{Training} and \textit{Validation} splits. In total, the MS COCO dataset contains 174,011 images. We used annotations provided by the COCO dataset to find and crop relevant object in each image. All images were padded by retaining an additional \texttildelow50\% of its cropped height and width (i.e. by dividing the height and width by $\sqrt{2}$ ). Padded image patches whose height or width was under 70 pixels were excluded to avoid extreme image transformations on later stages.
In total, we collected 55,388 object images that fall into the chosen twelve categories.  
We took all images from each of twelve categories with the exception of the ``person" category, which was reduced to 4,000 images in order to balance the overall number of images per category (the original ``person''  category has more than 120k images). See Figure~\ref{fig_cls_sample} for sample validation domain data. The breakdown of the validation domain dataset by the number of images per category is shown in Table~\ref{tab_num_of_models_and_images} (supplementary). 

\noindent \textbf{Testing Domain: YouTube Bounding Boxes.} Due to the overlap in object category labels with the other two domains, we chose the YouTube Bounding Boxes (YT-BB) dataset~\cite{real17youtube} to construct the test domain. Compared to the validation domain (MS COCO), the image resolution in YT-BB is much lower, because they are frames extracted from YouTube videos. The original YT-BB dataset contains segments extracted from 240,000 videos and approximately 5.6 million bounding box annotations for 23 categories of tracked objects. We extracted 72,372 frame crops that fall into one of our twelve categories and satisfy the size constraints. Some samples are shown in Figure~\ref{fig_cls_sample}.

\subsection{Experiments}
\label{sec_experimental_setup}
\noindent\textbf{Experimental Setup.} Our first set of experiments aims to provide a set of baselines for the VisDA-C challenge. We perform in-domain (\ie train and test on the same domain) experiments to obtain  approximate ``oracle'' performance, as well as source-only (\ie train only on the source domain) to obtain the lower bound results of no adaptation.
In total, we have 152,397 images as the source domain and 55,388 images as the target domain for validation. In our in-domain experiments, we follow a 70\%/30\% split for training and testing, i.e., 106,679 training images, 45,718 test images for the synthetic domain and 38,772 training images, 16,616 test images for the real domain. 

We first adopt the widely used AlexNet CNN architecture as the base model. The last layer is replaced with a fully connected layer with output size 12. We initialize the network with parameters learned on ImageNet~\cite{ILSVRC15}, except the for the last layer, which is initialized with random weights from $\mathcal{N}(0, 0.01)$. We utilize mini-batch stochastic gradient descent (SGD) and set the base learning rate to be $10^{-3}$, weight decay to be $5 \times 10^{-4}$ and momentum to be 0.9. We also utilize ResNext-152~\cite{resnext}; the output dimension of the last fully connected layer is changed to 12 and initialized with ``Xaiver'' parameter initializer~\cite{glorot2010understanding}. Since the output layer is trained from scratch, we set the learning rate to be 10 times that of other layers. The learning rate is adjusted with the formula: $\eta_{p} = \frac{\eta_{0}}{(1+\alpha p)^{\beta}}$, where $p$ will linearly change from 0 to 1 along the training process, $\eta_{0} = 10^{-4}$, $\alpha = 10$ and $\beta = 0.75$. We report the  accuracy of classification averaged over categories at 40k iterations.

\noindent\textbf{Domain Adaptation Algorithms.} We evaluate two existing domain adaptation algorithms as baselines. \textbf{\textit{DAN}} (Deep Adaptation Network)~\cite{long2015learning} learns transferable features by training deep models with Maximum Mean Discrepancy~\cite{sejdinovic2013equivalence} loss to align the feature distribution of source domain to target domain. In our implementation, the network architecture of \textbf{\textit{DAN}} is extended from AlexNet~\cite{alexnet}, which consists of 5 convolutional layers (\textit{conv1 - conv5}) and 3 fully connected layers (\textit{fc6 - fc8}) and \textbf{\textit{Deep CORAL}} (Deep Correlation Alignment)~\cite{sun2015return}  performs deep model adaptation by matching the second-order statistics of feature distributions. The domain discrepancy is then defined as the squared Frobenius norm $d(S, T) = \lVert \operatorname{Cov}_{S}-\operatorname{Cov}_{T}\rVert_{F}^2$, where $\operatorname{Cov}_{S}, \operatorname{Cov}_{T}$ are the covariance matrices of feature vectors from the source and target domain, respectively. 

\noindent\textbf{Baseline Results.} Baseline results on the validation domain for classification are shown in Table~\ref{tab_cls}. ``Oracle" or in-domain AlexNet performance for training and testing on the synthetic domain reaches 99.92\% accuracy, and training and testing on the real validation domain leads to 87.62\%. This supervised learning performance provides a loose upper bound for our adaptation algorithms. As far as unadapted source-only results on the validation dataset, AlexNet trained on the synthetic source domain and tested on the real domain obtains 28.12\% accuracy, a significant drop from in-domain performance. This provides a measure of how much the domain shift affects the model. Among the tested domain adaptation algorithms, Deep CORAL improves the cross-domain performance from 28.12\% to 45.53\% and DAN further boosts the result to 51.62\%. While their overall performance is not at the level of in-domain training, they achieve large relative improvements over the base model through unsupervised domain adaptation, improving it by 83.6\% and 61.9\% respectively.

In-domain oracle and source-only performance of AlexNet was similar on the test dataset to the validation dataset.  Oracle performance of AlexNet is 92.08\% and ResNext-152 improves the result to 93.40\%. Source AlexNet achieves 30.81\% mean accuracy, and DAN and Deep CORAL improve the result to 49.78\% and 45.29\%, respectively. As a base model, AlexNet has relatively low performance due to its simpler architecture, compared to more recent CNNs. However, the relative improvement of domain adaptation algorithms (\ie DAN and Deep CORAL) is still large.

\noindent\textbf{VisDA-C Challenge Results.} We provided the labeled source/validation data to challenge participants and set up an evaluation server for the test domain. The (unlabeled) test data was only made available several weeks before the end of the challenge.
Table~\ref{tab_cls} shows the top three of the  many submitted domain adaptation results. 

The top performing \textit{GFColourLabUEA} team used the label propagation algorithm \cite{french17self} that in turn was based on the recent $\Pi$-model for semi-supervised learning \cite{LaineA16} and the mean teacher model \cite{TarvainenV17}. It improved their source-only ResNet-152 model from 45.3\% to 92.8\%, a 104\% relative improvement. Their method consisted of optimizing two losses: 1) a mean cross entropy between ground truth and predictions of the so-called student network on samples from the source domain, and 2) a mean square difference between predictions of student and teacher networks on all samples from both domains. An essential step in the proposed method is to define a teacher network as having the same architecture as the student, but with its weights set to an \textit{exponential moving average of weights} of the student network on previous training iterations, therefore optimizing an agreement of the student network trained in a supervised fashion with its copies from the past on samples from both domains. Moreover, the resulting loss is trained in minibatches with dropout, noise and random data augmentations to improve robustness of the resulting procedure. More specifically, the ResNet-152 network pre-trained on Imagenet was used as a base model, its last fully connected classification layer was replaced by a sequence of randomly initialized fully connected layers, whereas the first convolutional layer and the layers leading up to the first downsampling were left as is. The rest of the network is first trained in a supervised fashion on source and then adapted on target using the method described above.

The second and third-best performing teams, \textit{NLE\_DA} and \textit{BUPT\_OVERFIT}, made use of various discrepancy measures such as maximum mean discrepancy (MMD) with different bandwidth priors measured between single resulting feature representation or combined representations from multiple layers (JMMD). Moreover, late fusion of several shallow networks built on features extracted from off the shelf deep models pretrained on Imagenet, and other recent approached to domain adaptation resulted in a substantial improvement in terms of the test score. 
We note that the top three teams had significantly higher source-only accuracy than our baselines, and some used model ensembling. While the top-1 team more than doubled their source performance, the two runner-ups had smaller relative improvements of 36\% and 35\%, respectively.

\subsection{Increasing Difficulty and Future Research}

\begin{figure}[t]
    \centering
    
    \includegraphics[width=.8\linewidth]{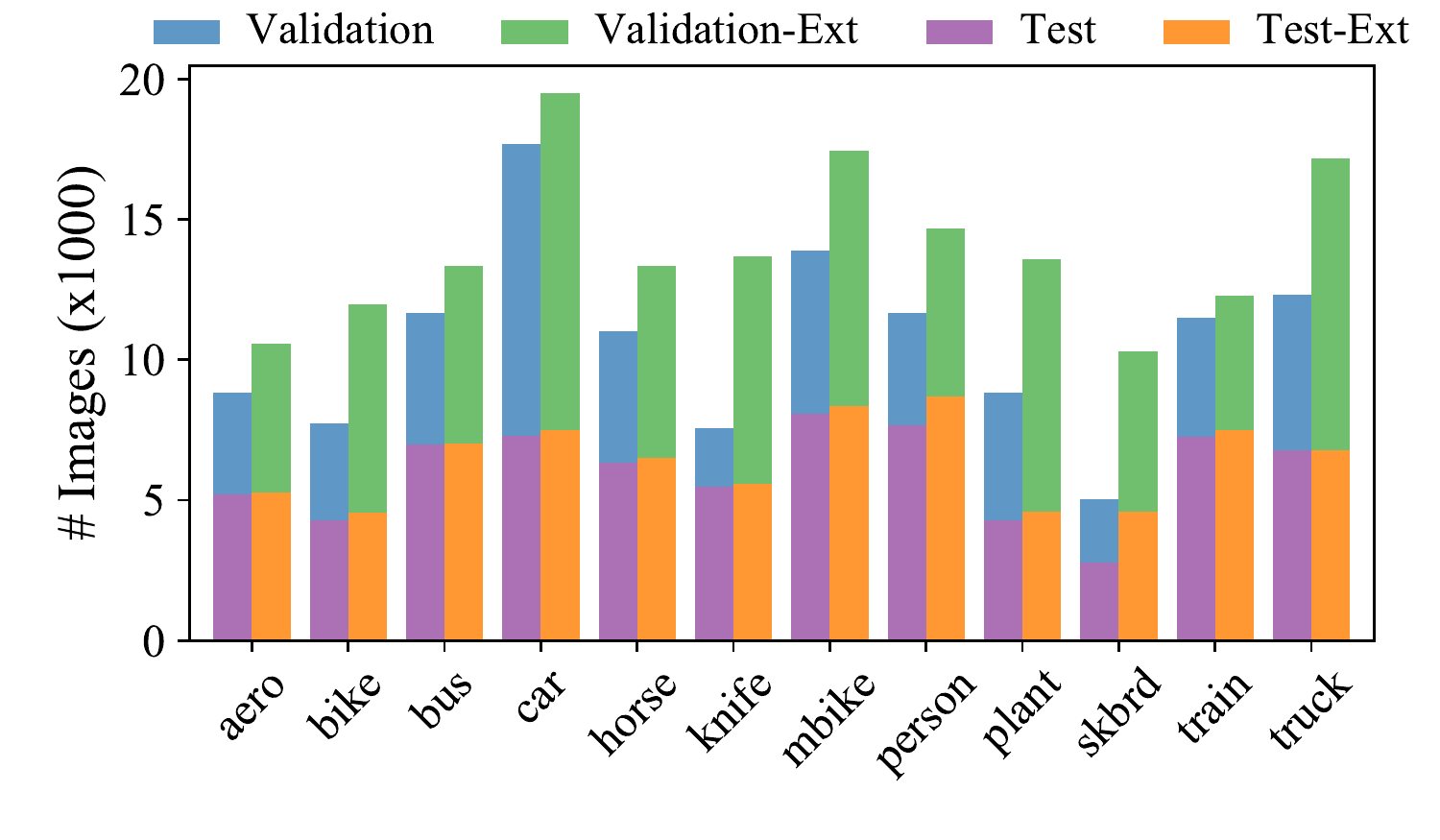}

    \vspace{-0.3cm}
    \caption{\small Comparison between VisDA-C and VisDA-C-ext in terms of per-category number of images. }
    \label{fig_cls_com_v1_v2}
    \centering
\end{figure}

\begin{figure}[t]
    \centering
    \includegraphics[trim={0 0 0 0.9cm},clip, width=\linewidth]{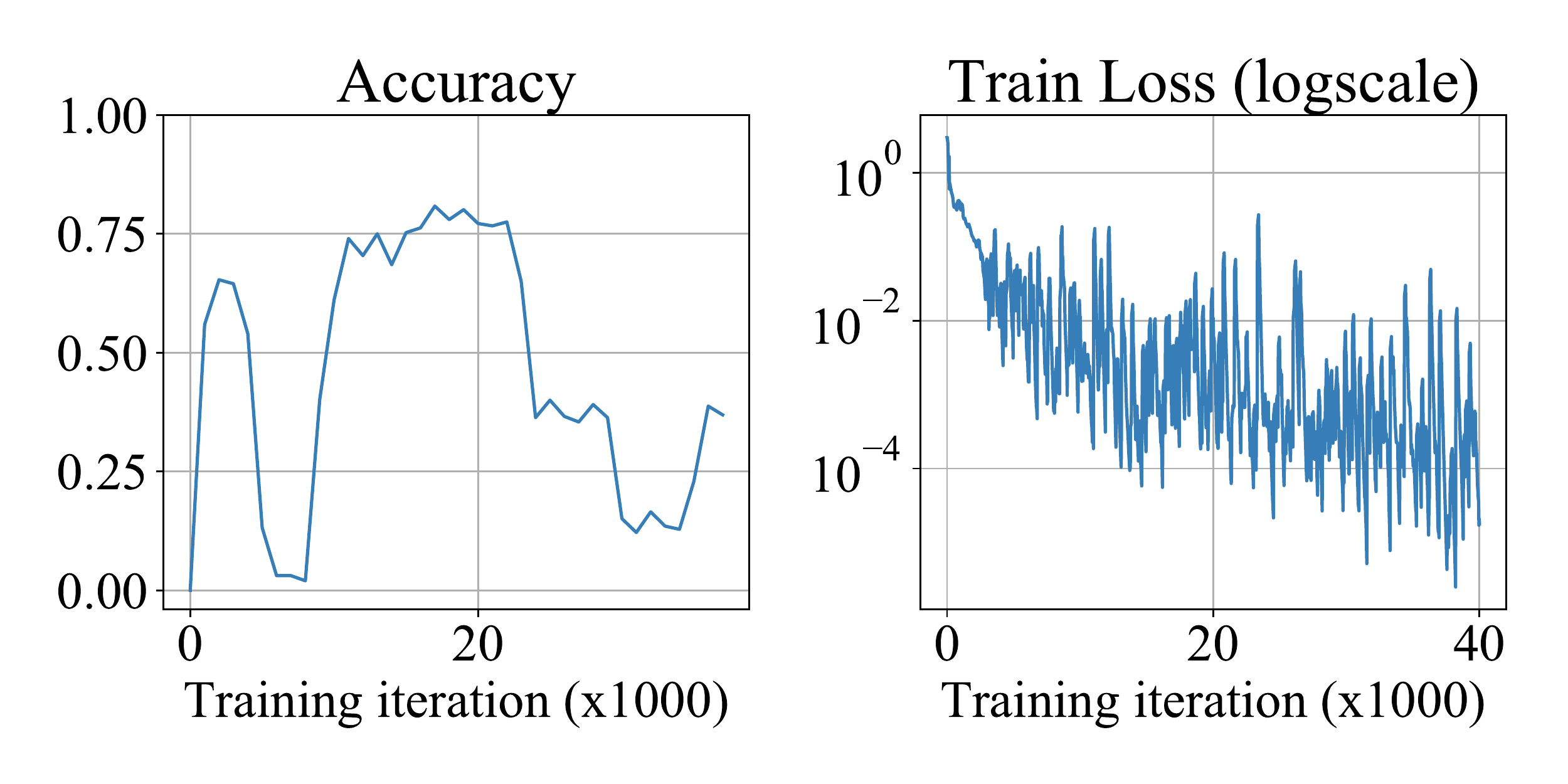} 
    \label{fig_analysis}
    \centering
    \vspace{-0.9cm}
\end{figure}

\begin{table}
\begin{center}
\small{
\begin{tabular}{ |c  !{\vrule width0.8pt} c | c| c | c| } 
\Xhline{0.8pt}
 \multirow{ 2}{*}{Method} &\multicolumn{2}{c}{VisDA-C}&\multicolumn{2}{c}{VisDA-C-ext}  \\
\cline {2-5}
 & val & test & val & test\\
\cline {1-5}
AlexNet~\cite{alexnet}  &28.12& 30.81 & 22.10 & 28.56  \\
ResNext-152~\cite{resnext} & 41.21 & 38.62 & 26.28  & 36.98 \\
ResNext-152+JMMD~\cite{JAN} & 64.54 & 58.37 & 47.68  & 54.73   \\
\Xhline{0.8pt}
\end{tabular}
} 
\end{center}
\vspace{-0.3cm}
\caption{\small Experimental results for VisDA-C and VisDA-C-ext obtained with different base models and a baseline adaptation method different from the one used in the Table \ref{tab_cls}.}
\label{tab_result_v2}
\vspace{-0.3cm}
\end{table}

The challenge participants obtained excellent performance in our test domain, on par with in-domain training. We therefore propose several ways to increase the difficulty of the task as a testbed for future research on this topic. 

\textbf{COCO $\leftrightarrow$ YTBB.} First, the test and validation domains can be swapped, as the MS-COCO domain turns out to be more difficult. We will shortly release ground truth labels for the test in order to enable this harder version of the task.

\textbf{VisDA-C Extended.} We also generate the VisDA-C-ext by adding smaller images to the real domains. In the initial version of the dataset, images whose height or width are under 70 pixels were excluded. We add these smaller or irregular-shaped images back to VisDA-C to increase its difficulty. Figure~\ref{fig_cls_com_v1_v2} shows the amount of data added in each category. In total, we add 35,591 images to the MS-COCO domain and 4,533 images to the YT-BB domain. 

To investigate how the newly added images affect the performance, we train three deep models (\ie AlexNet~\cite{alexnet}, ResNext-152~\cite{resnext} and ResNext-152+JMMD) on the VisDA-C training domain and test on both VisDA-C and VisDA-C-ext test domains. All the deep models are implemented in Caffe~\cite{caffe} and fine-tuned from models pre-trained on the ImageNet 2012 dataset. The experimental setup for AlexNet and ResNext-152 is the same with Section~\ref{sec_experimental_setup}. Joint Maximum Mean Discrepancy (JMMD)~\cite{JAN} is the distance between the means of kernelized representations of source and target samples, where sample representations are defined as \textit{combinations of outputs} of multiple layers of a network, therefore taking cross-terms between outputs of different layers into considerations. Inspired by~\cite{JAN}, we replaced the 1000-dimensional fully connected output layer with a sequence of 256-dimensional and 12-dimensional output layers (our dataset has 12 classes in total) and used combined outputs of these two layers as a feature representation.

The experimental results shown in Figure~\ref{tab_result_v2} indicate all three models get worse results on VisDA-C-ext, compared to VisDA-C. Moreover, the performance degradation on the validation domain (MS COCO) is larger than that of testing domain (YT-BB). The hyper-parameters of ResNext-152+JMMD are tuned on validation domain as we are not expected to use the labels in the testing domain. Figure~\ref{tab_result_v2} shows how the accuracy and training loss vary with iterations. The figure reveals that selecting the proper stopping criteria is important due to the drastic variation in reported accuracy. However, in the test domain, due to the lack of labels, there is no way to know how the accuracy will behave. Our conservative strategy is to choose the point where the first loss peak occurs (approximately 4,000 iterations). According to our strategy, ResNext-152+JMMD gets 64.54\% and 58.37\% mean accuracy on validation and test splits of VisDA-C respectfully. Hypothetically, our model can reach around 80\% accuracy on the test domain of VisDA-C if we were to ``cross validate on test servers'' and ensemble top performing models as some teams did during the challenge. 

\textbf{No Imagenet Pre-training.} Moreover, all participants used models trained in a supervised fashion on Imagenet as models to start from. Even though directly exploiting relations between Imagenet classes and classes used in our challenge was deliberately forbidden, the Imagenet pre-training made the resulting adaptation problem considerably easier, because resulting systems made extensive use of embedding that were designed to have high discriminative power in the first place. Considering that it is very hard to run domain adaptation synthetic-to-real challenges for domains that lack such enormous labeled datasets as Imagenet (e.g. medical imaging or decision making in robotics), it is even more important to design algorithms that can perform well \textit{without supervised pre-training} on similar domains.

\textbf{Variations in background/texture/POV.} We also release tools, models and required metadata to enable researchers to generate their versions of the dataset to increase its difficulty or to check specific hypothesis about domain adaptation, such as robustness of these methods to degrees of variations mentioned above, or to generate datasets for different tasks such as detection.

\section{VisDA-S: Semantic Segmentation}

The goal of our \textit{VisDA2017 Segmentation (VisDA-S)} benchmark is to test adaptation between synthetic  and real dashcam footage for semantic image segmentation. The training data includes pixel-level semantic annotations for 19 classes. We also provide validation and testing data, following same protocol as for classification:
    \vspace{-2mm}

\begin{itemize}
    \item \textbf{training domain (source):} synthetic dashcam renderings from the GTA5 dataset along with semantic segmentation labels,
    \vspace{-2mm}
    \item \textbf{validation domain (target):} a real-world collection of dashcam images from the CityScapes dataset along with semantic segmentation labels to be used for validating the unsupervised adaptation performance,
    \vspace{-2mm}
    \item \textbf{test domain (target):} a different set of unlabeled, real-world images from the new Nexar dashcam dataset.
    \vspace{-2mm}
    
\end{itemize}

\begin{table*}[tb]
\centering
\noindent\resizebox{\textwidth}{!}{\begin{tabular}{c|c c c c c c c c c c c c c c c c c c c c}
\multicolumn{21}{c}{\textit{\textbf{GTA $\rightarrow$ CityScapes Validation Domain}}}\\
\hline

 Method & {\rotatebox{90}{road}} & {\rotatebox{90}{sidewalk}} & {\rotatebox{90}{building}} & {\rotatebox{90}{wall}} & {\rotatebox{90}{fence}} & {\rotatebox{90}{pole}} & {\rotatebox{90}{t light}} & {\rotatebox{90}{t sign}} & {\rotatebox{90}{veg}} & {\rotatebox{90}{terrain}} & {\rotatebox{90}{sky}} & {\rotatebox{90}{person}} & {\rotatebox{90}{rider}} & {\rotatebox{90}{car}} & {\rotatebox{90}{truck}} & {\rotatebox{90}{bus}} & {\rotatebox{90}{train}} & {\rotatebox{90}{mbike}} & {\rotatebox{90}{bike}} & mIoU \\
\hline
Source (Dilation F.E.) & 30.6 &	21.2&	44.8&	10.1&	4.4&	15.4&	12.4&	1.7&	75.1&	13.5&	58.1&	38.0&	0.2&	67.5&	9.4&	5.0&	0.0&	0.0&	0.0&	21.4\\
Oracle (Dilation F.E.) & 96.2 &76.0& 88.4& 32.5& 46.4& 53.5& 52.0& 68.7& 88.6& 46.6& 91.0& 74.8& 46.0& 90.5& 46.9& 58.0& 44.7& 45.2& 70.3& 64.0\\

\hline
\multicolumn{21}{c}{\rule{0pt}{0.5cm}
\textbf{\textit{GTA $\rightarrow$ Nexar Test Domain}}}\\
\hline
Source (Dilation F.E.) & 40.7 & 19.2 & 42.3 & 4.2 & 20.0 & 21.8 & 26.0 & 13.4 & 68.0 & 19.6 & 84.7 & 32.4 & 5.8 & 59.0 & 10.3 & 9.8 & 1.6 & 13.5 & 0.0 & 25.9 \\
FCN-in-the-wild~\cite{hoffman2016fcns}  & 57.8 & 20.1 & 51.0 & 6.5 & 14.1 & 20.4 &  26.7 & 13.7 & 66.1 & 22.2 & 88.9 & 34.1 & 13.2 & 63.2 & 10.2 & 7.1 & 2.0 & 18.7 & 0.0 & 28.2  \\

\hline 

MSRA&	87.0&	38.5&	74.7&	23.7&	30.5&	41.1&	45.2&	36.9&	72.1&	32.6&	90.4&	55.9&	26.8&	80.0&	23.4&	25.1&	28.7&	44.6&	46.0&	\textbf{47.5} \\
Oxford &		85.3&	42.4&	53.4&	17.3&	31.5&	39.0&	45.6&	29.1&	77.1&	27.3&	61.7&	57.8&	46.1&	75.5&	27.5&	36.4&	26.3&	51.1&	18.1&	44.7 \\

VLLAB&	87.2&	 33.3&	70.2&	13.6&	27.8&	29.3&	32.9&	27.9&	77.2&	28.6&	90.3&	47.0&	35.7&	78.0&	24.8&	18.0&	9.1&	37.4&	38.1&	42.4 \\

\hline
\end{tabular}}

\caption{\small \textbf{Baseline and challenge results for the segmentation track}.
The top shows IoU results from \cite{hoffman2016fcns} for the source Dilation Front End model and its oracle performance on the validation CityScapes domain. \cite{hoffman2016fcns}'s adaptation methods achieves 27.1 mIoU. The bottom evaluates the same source and adapted models on the test Nexar domain, and shows results obtained by the top three challenge teams.}
\label{tab_seg}
\vspace{-0.3cm}
\end{table*}

\begin{figure}[t]
    \centering
    \includegraphics[width=\linewidth]{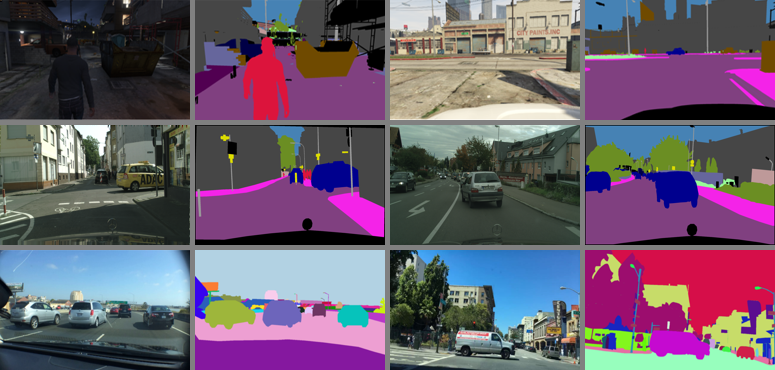}

    \caption{\small Images in the VisDA-S dataset. The first row shows the synthetic GTA5 images (training domain), the second row shows the images from CityScapes dataset (validation domain), the last row shows the images from Nexar dataset (test domain). }
    \label{fig_seg_sample}
    \centering
    \vspace{-5mm}
\end{figure}

The training and validation domain datasets used here are the same as those used in Hoffman et al (2016) \cite{hoffman2016fcns} for their work in synthetic to real adaptation in semantic segmentation tasks.

\noindent \textbf{Training Domain: Synthetic GTA5.} The images in the segmentation training come from the GTA5 dataset. GTA5 consists of 24,966 high quality labeled frames from the photorealistic, open-world computer game, Grand Theft Auto~V (GTA5). The frames are synthesized from a fictional city modeled off of Los Angeles, CA and are in high-resolution, $1914 \times $1052. All semantic segmentation labels used in the GTA5 dataset have a counterpart in the CityScapes category list for adaptation. See Figure \ref{fig_seg_sample} for sample training domain data.

\noindent \textbf{Validation Domain: Real CityScapes.} Data in the segmentation validation domain comes from the CityScapes dataset. CityScapes contains 5,000 dashcam photos separated by the individual European cities from which they were taken, with a breakdown of 2,975 training, 500 validation and 1,525 test images. Images are in high resolution, $2048 \times 1024$. In total, the CityScapes dataset has 34 semantic segmentation categories, of which we are interested in the 19 that overlap with the synthetic GTA5 dataset. See Figure \ref{fig_seg_sample} for sample validation domain data.

\noindent \textbf{Test Domain: Real DashCam Images. } Dashcam photos in the test domain were taken from a dataset recently released by Berkeley Deep Drive and Nexar~\cite{nexet}. They were collected using the Nexar dashcam interface and manually annotated with segmentation labels. We use 1500 images of size $1280\times720$ available with annotations corresponding to the 19 categories matching GTA5 and CityScapes. Note that this data along with the annotations is part of a larger data collection effort by Berkeley Deep Drive (BDD). See Figure~\ref{fig_seg_sample} for sample test domain data.

\noindent\textbf{Domain Adaptation Algorithms.} For details on the domain adaptation algorithms applied to this domain shift, we refer the reader to the original work that performed adaptation from GTA5 (synthetic) to CityScapes (real) in~\cite{hoffman2016fcns}. The authors use the front-end dilated fully convolutional network as the baseline model. The method for domain adaptive semantic segmentation consists of both global and category specific adaptation techniques. Please see section 3 (Fully Convolutional Adaptation Models) in~\cite{hoffman2016fcns} for detailed information about these techniques and their implementation. In all experiments, the Intersection over Union (IoU) evaluation metric is used to determine per-category segmentation performance. 

\noindent\textbf{Baseline Results.} Please refer to Table~\ref{tab_seg} and Section 4.2.1 in Hoffman\textit{ et al}.~\cite{hoffman2016fcns} for full experimental results and discussion of semantic segmentation performance in GTA5$\rightarrow$CityScapes adaptation. Some relevant results are replicated here. In summary, the front-end dilation source achieves a mean IoU (mIoU) of 21.6 over all semantic categories on the val domain, compared to oracle mIoU of 64.0. The adaptation method in~\cite{hoffman2016fcns} improves mIoU to 25.5. A similar performance improvement is seen when adapting the GTA5 model to our challenge test domain. 

\noindent\textbf{Challenge Results.} The winning team used a multi-stage procedure to improve performance of their adaptation pipeline. On the first stage, they trained and applied frame-level discriminator and updated pixels of target images in a way that preserves high level semantic features of objects, but makes each target frame as a whole less distinguishable from source frames. This approach resembles style-transfer initially proposed several years ago \cite{GatysEB15a}, but significantly improved since then \cite{JohnsonAL16, UlyanovLVL16, CycleGAN2017, LiW16b}. To perform second-stage pixel level discrimination winning team ensembled ResNet-101, ResNet-152 and SE-ResNeXt-101 with pyramid spatial pooling to obtain better feature representations and applied domain discrimination with multiple scaling parameters to get a more powerful yet stable discriminator. Other top performing teams used single-step procedures with pixel-level loss and different base models, such as DeepLab-v2, with smart balancing between weights of segmentation and adversarial losses.
\vspace{0.2cm}
\section{Conclusion}

\vspace{0.2cm}

In this paper, we introduce a large scale synthetic-to-real dataset for unsupervised domain adaptation, and present a detailed analysis and current state of the art performance.
We also describe several modifications to the dataset that increase its difficulty as well as alternative setups that enable verification of more complex hypotheses regarding domain adaptation methods. We highly encourage researchers to work on adaptation methods that \textit{do not rely on the supervised pre-training}, because there are plenty of important domains, such as medical imaging, that seriously lack labeled data, and therefore might greatly benefit from synthetic-to-real domain adaptation. Large scale synthetic-to-real datasets as the one described in this paper present an experimental setup designed for figuring out how to train these models without supervised pre-training, and therefore working on methods that perform well in practical domains that do not have large labeled datasets using simulated data, which is becoming more and more important these days. As of now the no-pre-train setup poses a substantial challenge for existing domain adaptation methods and solving it would greatly benefit the society.

\vspace{0.1cm}
Our plan for the nearest future is to opensource and support all resources, tools and dataset flavours discussed in this paper to enable the community to improve adaptation methods' performance in diverse and not yet considered situations. We hope that this benchmark will  be used by the wider research community to develop and test novel domain adaptation models using an established protocol.

\clearpage
{\small
\bibliographystyle{ieee}
\bibliography{egbib}
}

\clearpage

\section{Supplementary Appendix}
\renewcommand{\thesubsection}{\Alph{subsection}}

\subsection{Other relevant datasets}
In table~\ref{tab:synth_datasets}, we listed the existing cross-domain datasets used for object classification and semantic segmentation. Other relevant datasets are showed in Table~\ref{fig:datasets_extra}.
\begin{table}[h]
    \centering
    DIGIT CLASSIFICATION
    \begin{tabular}{|c|c c c|}\hline
    Dataset & Examples & Classes & Domains \\ \hline
    USPS-subset \cite{hull1994database} & 1,800     & 10& 1 \\
    MNIST-subset \cite{lecun1998gradient} & 2,000     & 10& 1 \\
    USPS-Full \cite{hull1994database}       & 9,000     & 10& 1 \\
    MNIST-Full \cite{lecun1998gradient}  & 70,000    & 10& 1 \\
    SVHN \cite{netzer2011reading} & 630,420   & 10& 1 \\ \hline
    \end{tabular}
~\\    
    FACE RECOGNITION \\
    \begin{tabular}{|c|c c c|} \hline
    Dataset & Examples & Classes & Domains \\ \hline
    PIE \cite{sim2002cmu}  & 11,554    & 68& 1 \\ \hline 
    \end{tabular} \\
~\\    
    OBJECT DETECTION \\
    \small{
    \begin{tabular}{|c|c c c|} \hline
    Dataset & Examples & Classes & Domains \\ \hline
    CAD-PASCAL~\cite{peng2015learning} & 12,000 & 20& 2 \\    
    NEXAR \cite{nexet} & 55,000 & 1& 3\\ 
    TPAMI14 \cite{vazquez2014virtual} & 5,616 & 1& 2 \\ \hline
    \end{tabular}} \\
    \caption{Other popular domain adaptation datasets that are not directly comparable with VisDA, because of either non-real domain (digits, faces) or different tasks (detection). However, note that these datasets are either smaller or are limited in terms of domain coverage.}
    \label{fig:datasets_extra}
\end{table}

\subsection{Number of images per category in VisDA-C}
Table~\ref{tab_num_of_models_and_images} shows number of models and images in VisDA-C in detail.

\begin{table}[h]
\begin{center}
\begin{tabular}{ c | c c| c | c } 
\hline
Category  &\multicolumn{2}{c}{Training}&\multicolumn{1}{c}{Validation} & Testing \\
\hline
& Models  & Images & Images & Images\\
\cline {2-5}
\rule{0pt}{2ex}    
aeroplane &179& 14,309 & 3,646 & 5,196  \\
bicycle &93& 7,365 & 3,475  &4,272 \\
bus &208& 16,640 & 4,690   &6,970  \\
car &160& 12,800 & 10,401   &7,280  \\
horse &119& 9,512 & 4,691  &6,331 \\
knife &178& 14,240 & 2,075  &5,491\\
motorcycle &217& 17,360 & 5,796&8,079\\
person &152& 12,160 & 4,000 &7,673 \\
plant &135& 10,371 & 4,549 &4,287 \\
skateboard &146& 11,680 & 2,281 &2,762\\
train &200& 16,000 & 4,236  &7,264\\
truck &120 & 9,600 & 5,548  &6,767\\
\hline
\rule{0pt}{2ex}    
total &1,907& 152,397 & 55,388  & 72,372\\
\hline
\end{tabular}
\end{center}
\vspace{-0.2cm}
\caption{ Number of models and images per category in VisDA Classification training, validation and testing domains. The image disrtibution is also shown as a bar plot in Figure \ref{pie_num_of_models_and_images}.} 
\vspace{-0.3cm}
\label{tab_num_of_models_and_images}
\end{table}

\subsection{VisDA Image samples}

\begin{figure*}
\centering
\includegraphics[width= 0.9\linewidth]{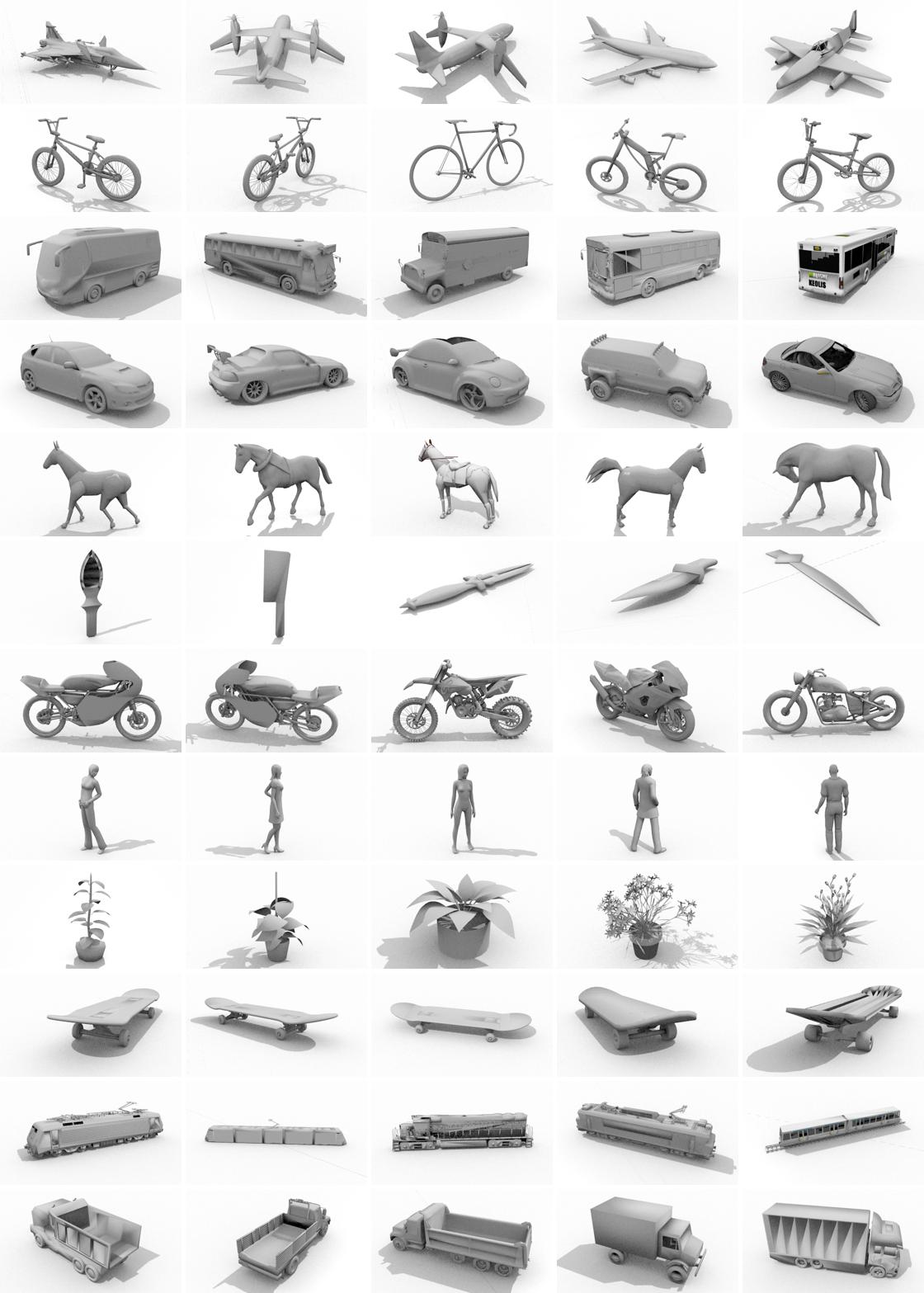}
\caption{\textbf{Classification training domain sample data.} Synthetic CAD models.}
\label{class_train_data}
\end{figure*}

\begin{figure*}
\centering
\includegraphics[width= 0.9\linewidth]{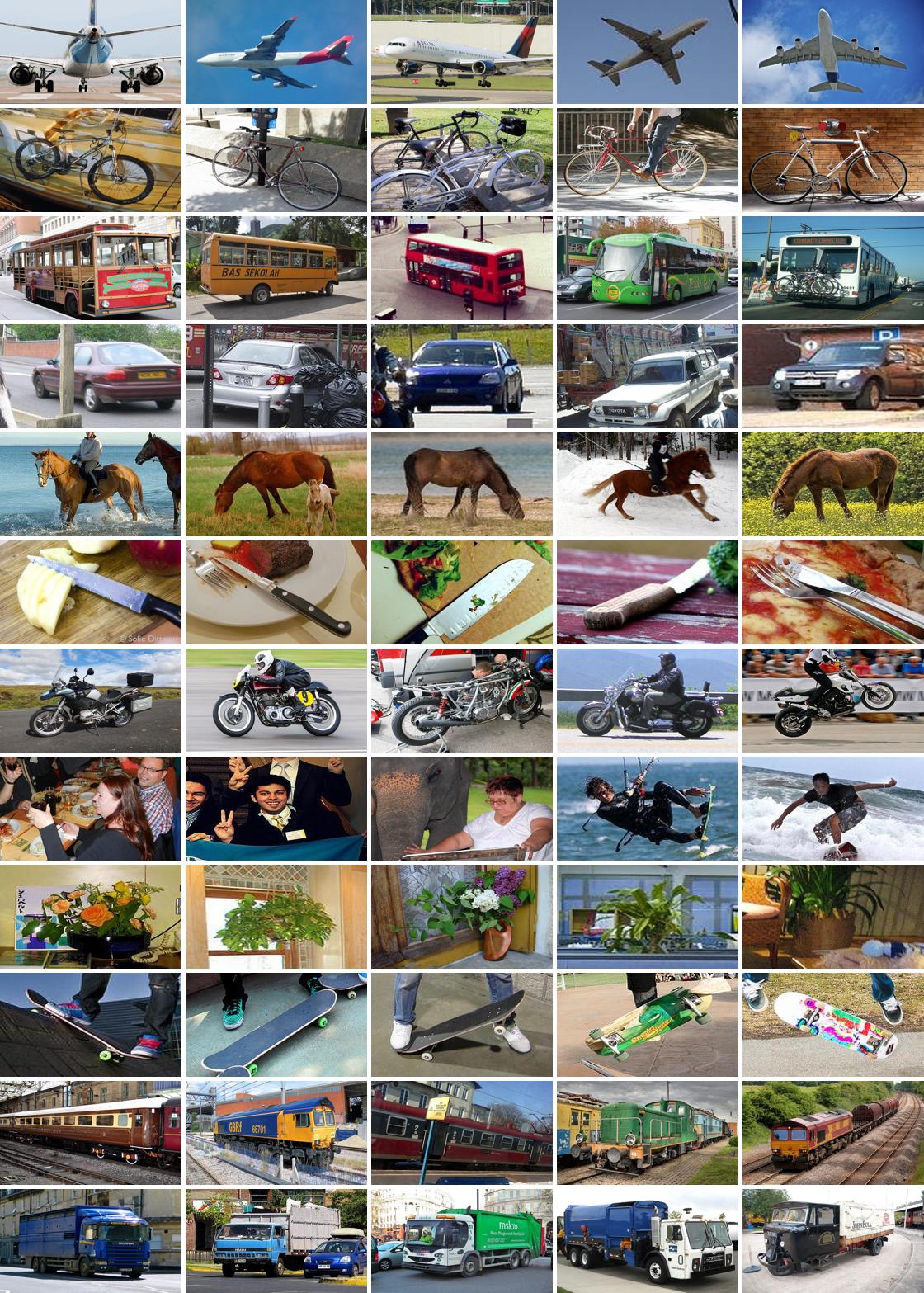}
\caption{\textbf{Classification validation domain sample data.} Real images from the MS COCO dataset.}
\label{class_val_data}
\end{figure*}

\begin{figure*}
\centering
\includegraphics[width= 0.9 \linewidth]{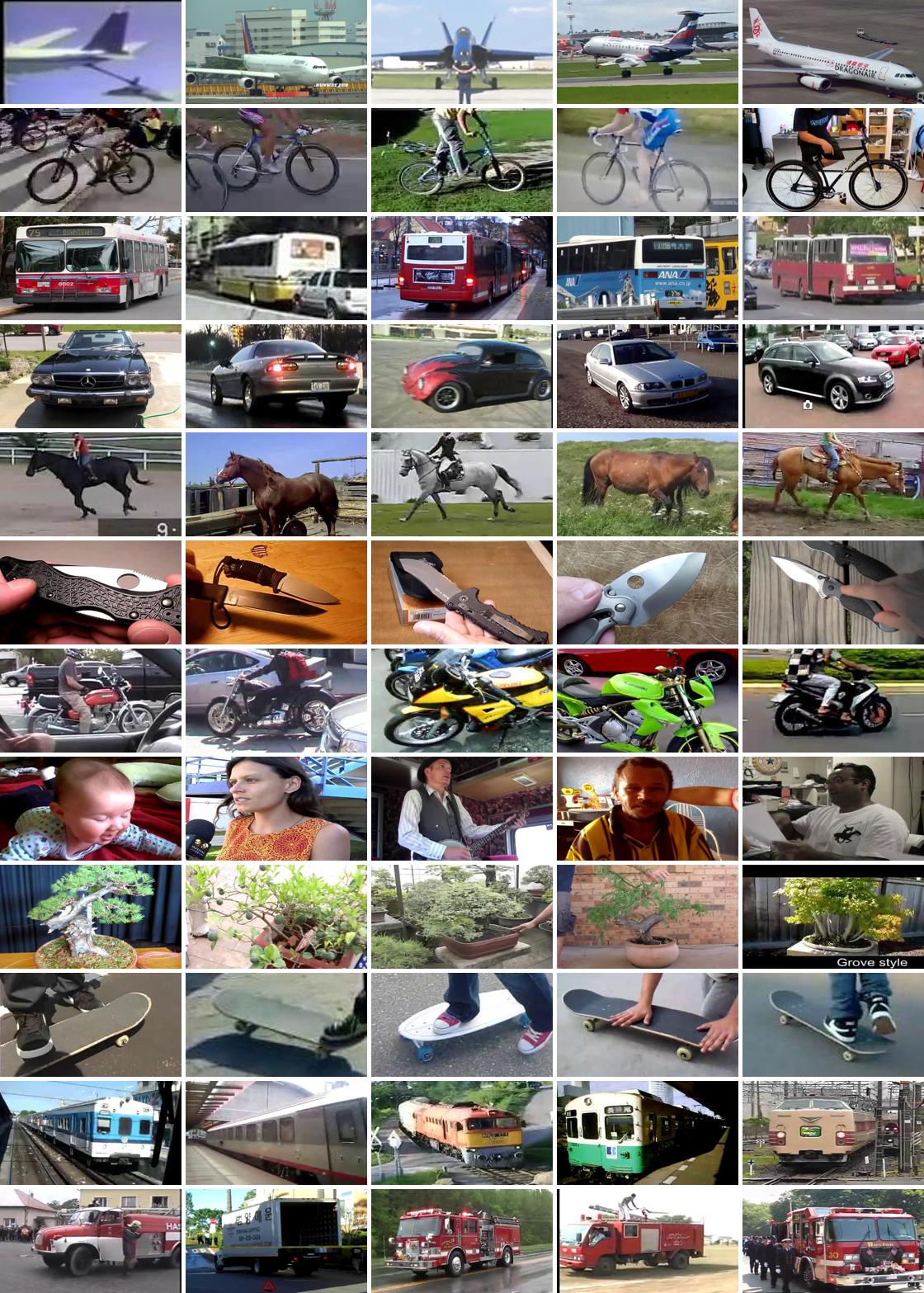}
\caption{\textbf{Classification test domain sample data.} Real images from YouTube-BB dataset.}
\label{class_test_data}
\end{figure*}

\begin{figure*}
\centering
\includegraphics[width= 0.9\linewidth]{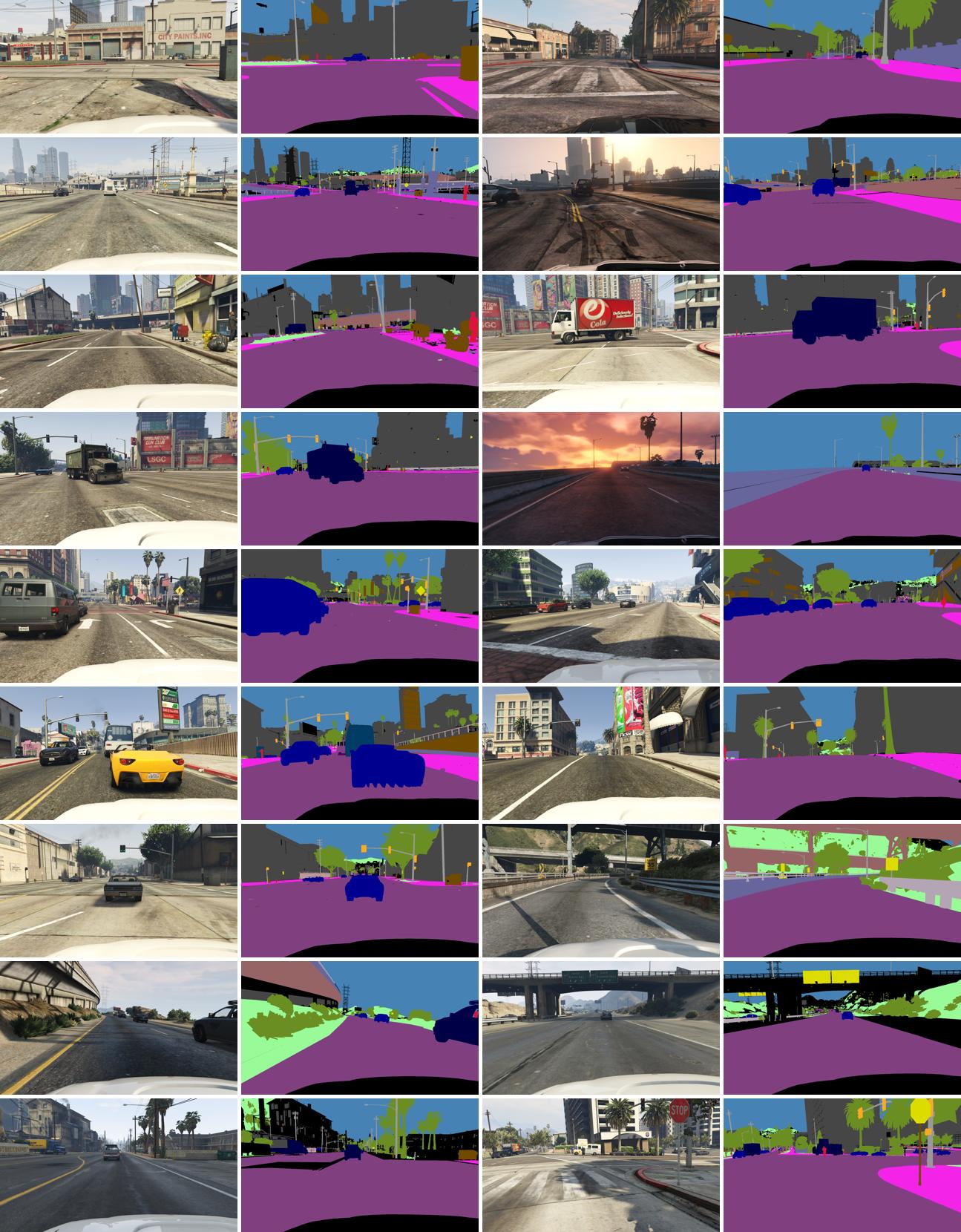}
\caption{\textbf{Segmentation training domain sample data.} Synthetic dashcam images from the GTA5 dataset.}
\label{seg_train_data}
\end{figure*}

\begin{figure*}
\centering
\includegraphics[width= 0.9\linewidth]{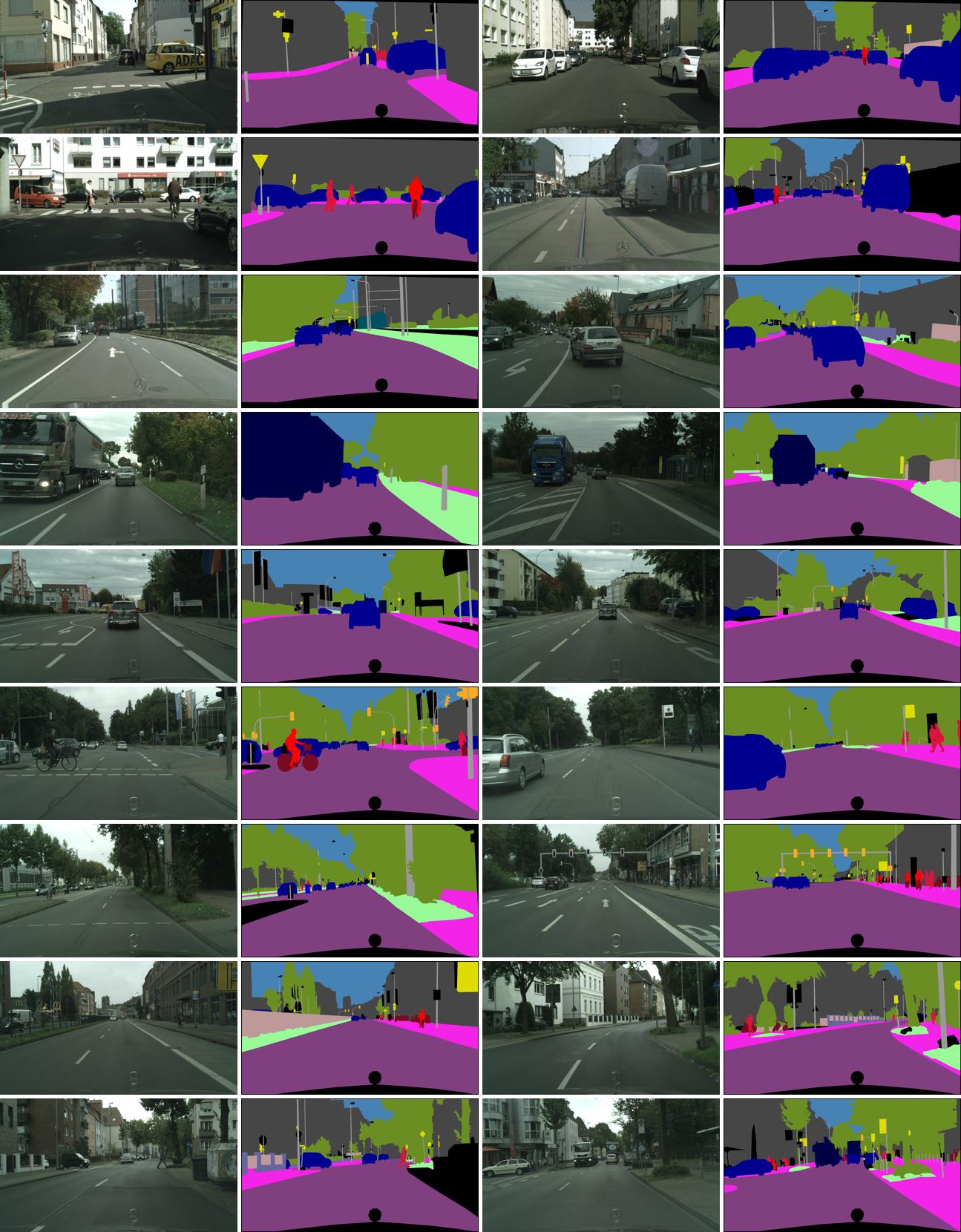}
\caption{\textbf{Segmentation validation domain sample data.} Real dashcam images from the CityScapes dataset.}
\label{seg_val_data}
\end{figure*}

\begin{figure*}
\centering
\includegraphics[width= 0.9\linewidth]{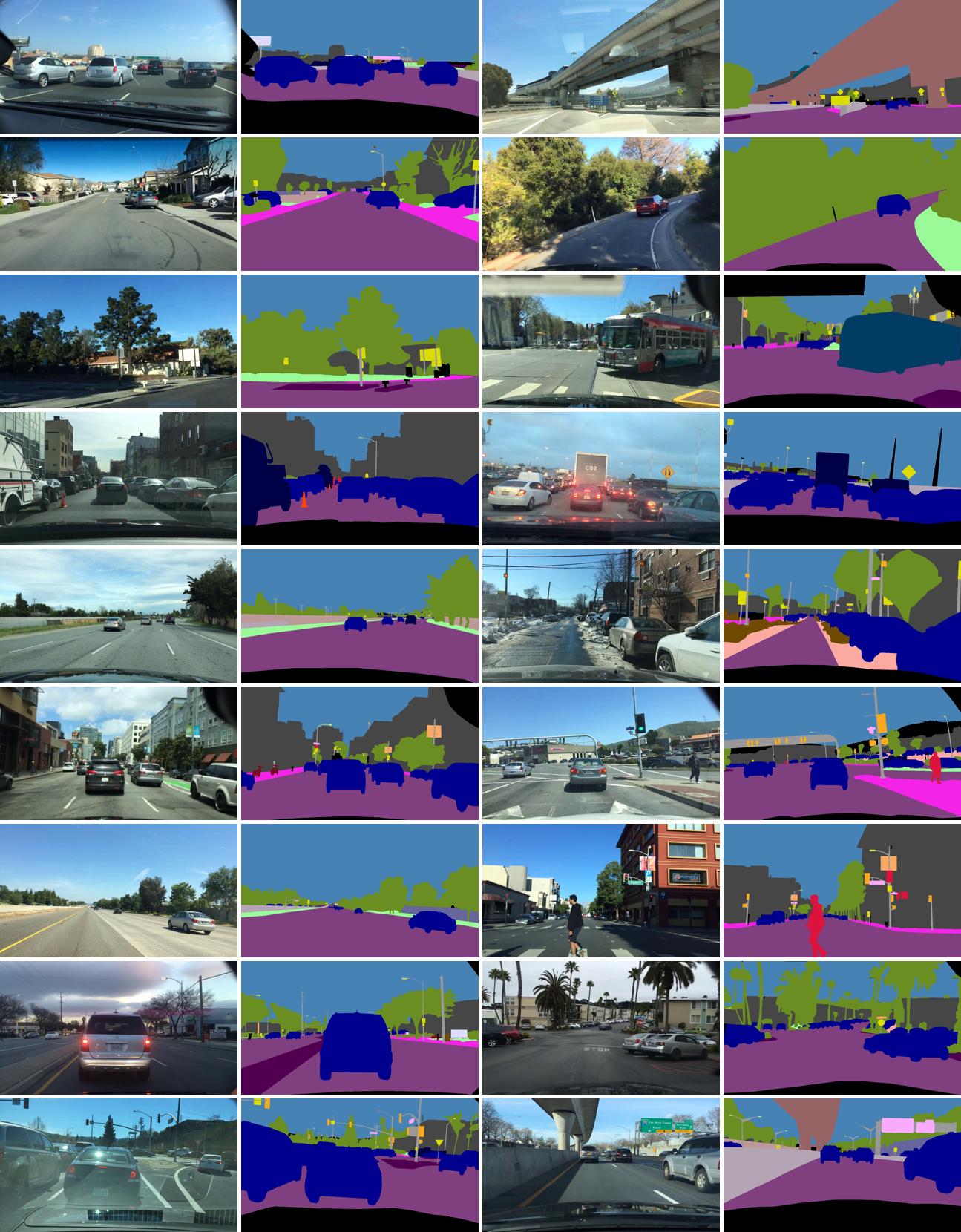}
\caption{\textbf{Segmentation test domain sample data.} Real dashcam images from the  NEXAR dataset.}
\label{seg_test_data}
\end{figure*}

We presented images from VisDA-C and VisDA-S in Figures~\ref{fig_cls_sample} and \ref{fig_seg_sample}, respectively. More images of each considered category for all domains are given in Figures~\ref{class_train_data}--\ref{seg_test_data}.

\clearpage

\clearpage
\clearpage

\end{document}